\documentclass{article}
\usepackage{ijcai23}

\usepackage{times}
\usepackage{soul}
\usepackage{url}
\usepackage[hidelinks]{hyperref}
\usepackage[utf8]{inputenc}
\usepackage[small]{caption}
\usepackage{graphicx}
\usepackage{amsmath}
\usepackage{amsthm}
\usepackage{booktabs}
\usepackage{algorithm}
\usepackage{algorithmic}
\usepackage[switch]{lineno}

\usepackage{microtype}
\usepackage{subfigure}
\usepackage{amsfonts,amssymb}
\usepackage[svgnames]{xcolor}
\usepackage{bm}
\usepackage{stfloats}

\title{Non-zero-sum Game Control for Multi-vehicle\\
Driving via Reinforcement Learning}

\author{
Xujie Song$^1$
\And
Zexi Lin$^2$
\affiliations
$^1$School of Vehicle and Mobility, Tsinghua University, Beijing, China\\
$^2$Department of Computer Science and Technology, Tsinghua University, Beijing, China
\emails
\{songxj21, linzx22\}@mails.tsinghua.edu.cn
}

\begin{document}

\maketitle

\begin{abstract}
    When a vehicle drives on the road, its behaviors will be affected by surrounding vehicles. 
    Prediction and decision should not be considered as two separate stages because all vehicles make decisions interactively.
    This paper constructs the multi-vehicle driving scenario as a non-zero-sum game and proposes a novel game control framework,
    which consider prediction, decision and control as a whole.
    The mutual influence of interactions between vehicles is considered in this framework because decisions are made by Nash equilibrium strategy.
    To efficiently obtain the strategy, ADP, a model-based reinforcement learning method, is used to solve coupled Hamilton-Jacobi-Bellman equations.
    Driving performance is evaluated by tracking, efficiency, safety and comfort indices.
    Experiments show that our algorithm could drive perfectly by directly controlling acceleration and steering angle.
    Vehicles could learn interactive behaviors such as overtaking and pass.
    In summary, we propose a non-zero-sum game framework for modeling multi-vehicle driving,
    provide an effective way to solve the Nash equilibrium driving strategy, and validate at non-signalized intersections.
\end{abstract}

\section{Introduction}

Multi-vehicles driving on the road is a complex interactive game problem \cite{xiaoming2010study,li2006cooperative}.
It requires each vehicle to decide driving strategy considering surrounding vehicles in real-time.
Therefore, it is necessary to design a driving control model which can make decisions interactively.

With the rapid development of artificial intelligence and V2X technologies, autonomous driving has gradually become possible.
Autonomous vehicles driving on the road also faces game problems \cite{dreves2018generalized}.
In traditional planning and decision-making methods, the trajectories of surrounding vehicles are predicted at first and the predicted trajectories are input as obstacles into the planning and decision-making module \cite{grigorescu2020survey}.
This two-stages method does not consider the influence of interactions between the ego car and surrounding cars.
In order to deal with this limitation, we absorb game theory into decision-making model to consider the mutual influence of the ego car and surrounding cars.
\begin{figure}[t]
    \begin{center}
    \centerline{\includegraphics[width=0.8\columnwidth]{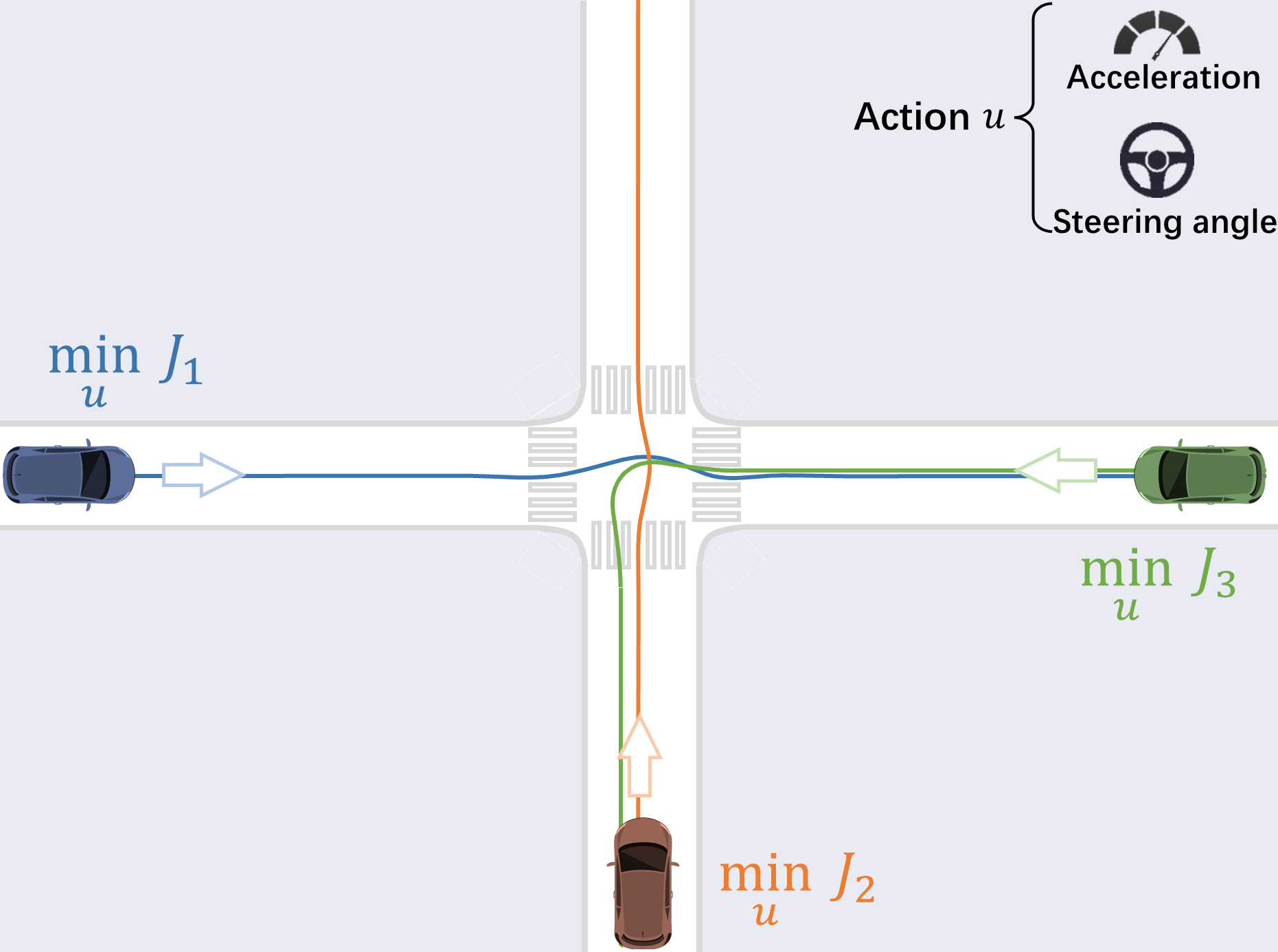}}
    \caption{\textbf{Three vehicles drive at a non-signalized intersection}.
             They have individual objective functions $J_i$, which involves their destinations, the reference trajectories, the expected speeds, the acceptable safe distances, et al.
             The moving of vehicles is controlled by acceleration and steering angle.
             This multi-player decision-making scenario with individual objective functions is a non-zero-sum game problem.}
    \label{intersection_scenario}
    \end{center}
\end{figure}

Several scholars have conducted in-depth research on driving games, but there are still some problems.
In 2015, game theory is introduced in intersection control for the first time \cite{elhenawy2015intersection}.
It was found that the traffic efficiency was improved by 49\% compared to using stop signs.
\cite{yang2016cooperative} and \cite{cheng2018speed} use Pareto optimality, a global optimal state, as the optimal driving control strategy.
However, vehicles tend to be non-cooperative in the real world, so it is unrealistic to reach the global optimal state.
To address this limitation, \cite{cheng2019vehicle} used Nash equilibrium strategy as the optimal driving strategy to achieve individual's optimality.
However, they assume that vehicles can only be controlled by acceleration without steering angle. It makes vehicles must travel on fixed trajectories regardless of trajectory offset.

Moreover, the above methods tend to define the range of acceleration as a discrete finite set, for example, \{$0$ m/s, $\pm1$ m/s, $\pm2$ m/s\}.
In reality, acceleration should be a continuous variable.
There are some works that partially solve those problems.
For example, \cite{fridovich2020iterative} uses the iterative quadratic method to solve a general-sum driving game.
However, they require the system dynamics be linearizable, which is unrealistic in complex driving games with highly non-linear vehicle dynamics.
\cite{cleac2019algames} proposes ALGAMES, which is an augmented Lagrangian game-theoretic solver for driving games.
They use MPC (Model Predictive Control) to solve a constrained game problem based on optimization methods.
However, when the number of cars becomes large, optimization-based methods will take a long time to find the optimal solution.
The large time usage makes it impossible in real-time decision-making and control.

To overcome those issues mentioned above, we adopt the 3-DOF (Degree of Freedom) vehicle dynamic model \cite{ge2021numerically} rather than the vehicle kinematic model,
to achieve joint control of acceleration and steering angle.
The control variables are both defined in continuous space for realistic.
To achieve better driving performance, we consider long-term cost based on tracking performance, efficiency, safety and comfort index in the future.
Then, multi-vehicles driving on the road is modeled as a non-zero-sum game problem.
To obtain high-precision control actions, we directly solve the original nonlinear problem rather than linearizing it like \cite{fridovich2020iterative}.
Finally, to solve the game problems and implement in real-time applications, we adopt the model-based reinforcement learning method, ADP (Approximate Dynamic Programming), to learn a neural network driving control policy.

\textbf{Contributions. } We summarize our key contributions as follows:
\begin{itemize}
    \item We propose a non-zero-sum game framework for modeling multi-vehicles driving.
          This framework can consider the influence of interactions between cars for improving driving intelligence.
    \item We provide an effective way to solve the Nash equilibrium driving strategy based on reinforcement learning.
          Experience replay and self-play tricks is adopted for scaling up the number of agents.
    \item We provide experiments showing our algorithm could complete trajectory tracking, speed tracking and collision avoidance task perfectly.
          Vehicles can learn to overtake and pass behaviors without prior knowledge.
    \item We release our code to facilitate reproducibility and future research \footnote{{\rm https://github.com/jerry99s/NonZeroSum\_Driving}}. 
\end{itemize}

\section{Background}
\subsection{Game Theory}

Game theory is an effective method to study the interactive control of multi-agents.
It studies how individuals obtain their best strategies while facing conflicts \cite{von1947theory}.
It has been applied to several driving control problems, such as lane changing control \cite{wang2015game,talebpour2015modeling} and driver-automation shared control \cite{na2014game,flad2014steering}.

According to the cooperative and competitive relationship between participants,
game problems can be divided into zero-sum games, non-zero-sum games, and completely cooperative games.
The participants in a zero-sum game try to maximize or minimize a same objective function,
while they have individual objective functions in a non-zero-sum game.
For example, AlphaGo \cite{silver2016mastering} models Go as a zero-sum game,
while Pluribus \cite{brown2019superhuman} models Texas hold'em poker as a non-zero-sum game.

\subsection{Nash Equilibrium}

\textbf{Definition 2.1} \cite{nash1950equilibrium} Let $u_i$ and $J_i(u_1,...,u_n)$ be the action and objective function of participant $i$.
The \textit{Nash equilibrium} point $(u_1^*,...,u_n^*)$ satisfies:
\[J_i(u_1^*,...,u_i^*,...,u_n^*) \leq J_i(u_1^*,...,u_i,...,u_n^*), \forall u_i \forall i\]
\begin{figure}[H]
    \begin{center}
    \centerline{\includegraphics[width=0.9\columnwidth]{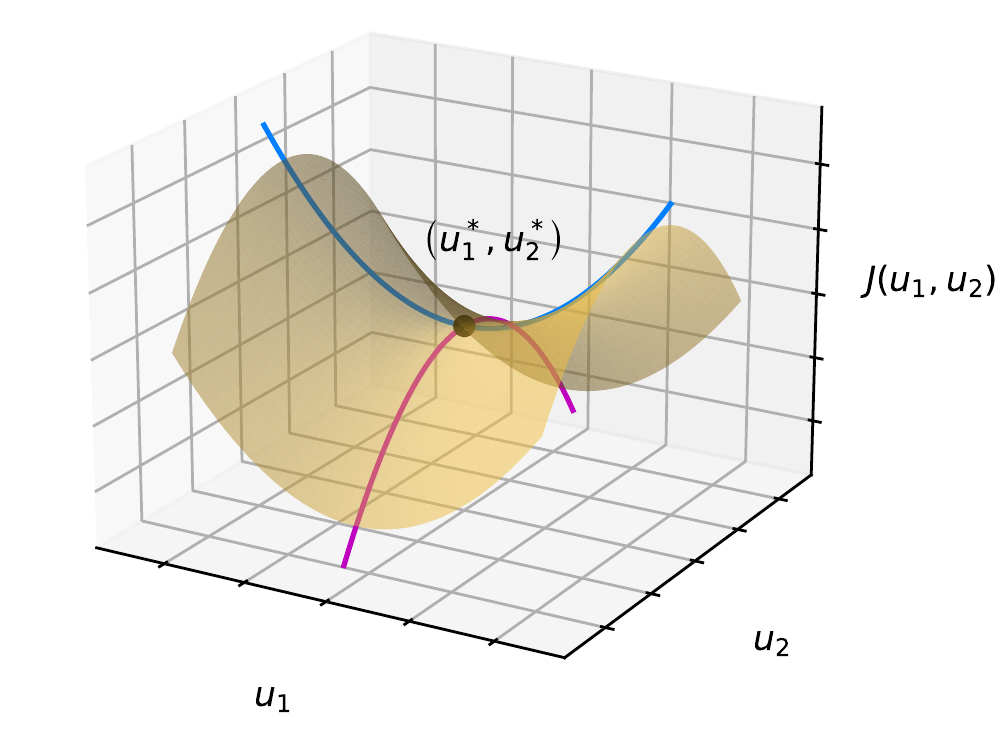}}
    \caption{\textbf{Nash equilibrium for a two-player zero-sum game.}
             Two players have the same objective function $J(u_1,u_2)$
             while player 1 tries to minimize it and player 2 tries to maximize it.
             They reach the Nash equilibrium at $(u_1^*, u_2^*)$.}
    \end{center}
\end{figure}
Nash equilibrium is an important concept in game theory. It describes the equilibrium state in non-cooperative games.
At the Nash equilibrium point, it is impossible for every participant to obtain a lower objective function value by only changing his own action.
Therefore, the Nash equilibrium action $u^*$ is the optimal action for each individuals if we suppose that all players are completely rational.
It inspires that we can find the Nash equilibrium strategy , which always give the Nash equilibrium action at whatever state, to improve individual's intelligence.
So, in the non-zero-sum multi-vehicles driving games described in Figure \ref{intersection_scenario}, we aims to find the Nash equilibrium driving strategies.

\subsection{Approximate Dynamic Programming}
\begin{figure}[H]
    \begin{center}
    \centerline{\includegraphics[width=0.8\columnwidth]{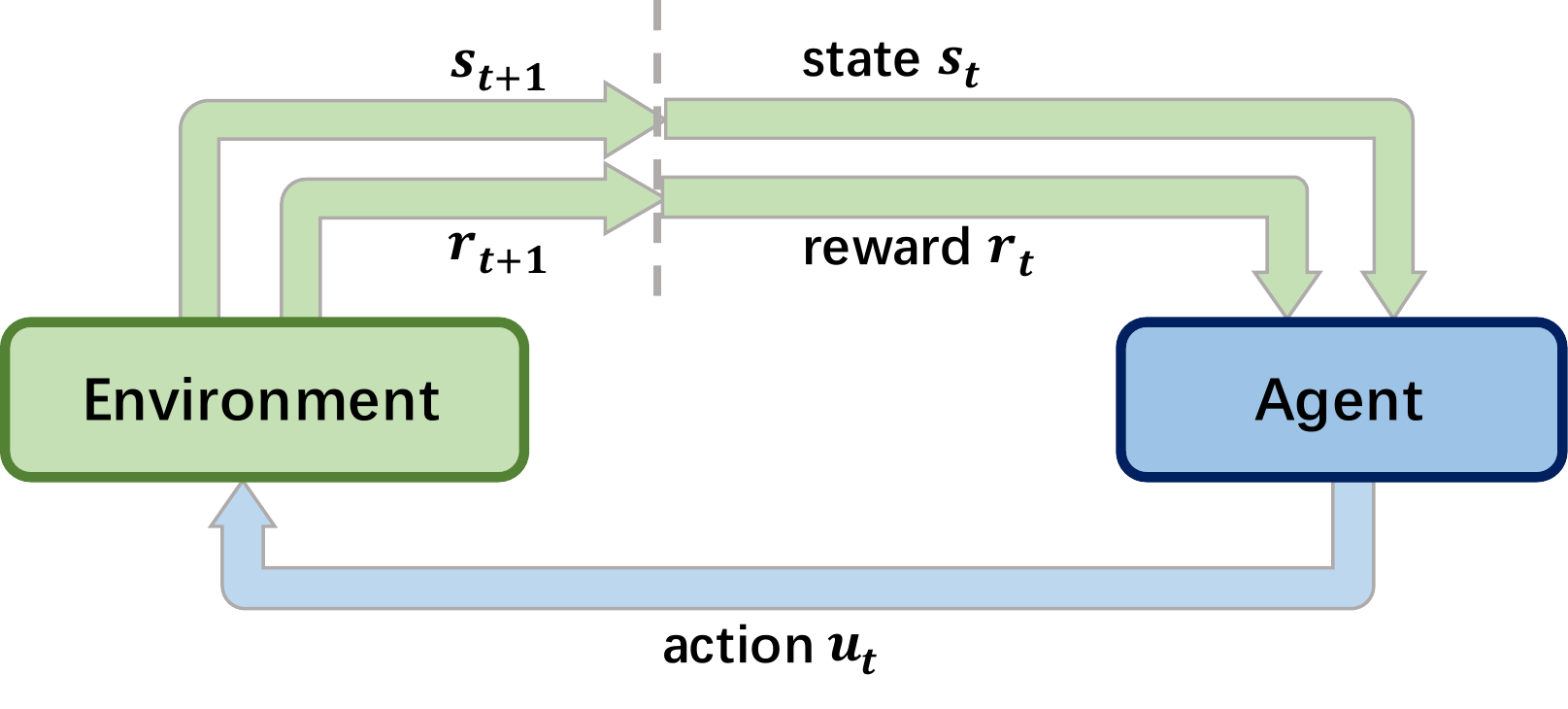}}
    \caption{\textbf{Reinforcement learning}: It perceives the current state and takes action to interact with the environment.
    Environment will update the state according to action and give reward or punishment feedback for action evaluation.}
    \label{RL_learning}
    \end{center}
\end{figure}

Reinforcement learning is a learning method by imitating the learning process of human beings
via "trial and error-evaluation".
The learning process is shown in Figure \ref{RL_learning}.
According to whether the learning method uses an environment dynamic model,
reinforcement learning can be divided into model-based category and model-free category.
ADP (Approximate dynamic programming) is a model-based reinforcement learning method \cite{lewis2009reinforcement}.

ADP learns a actor network $\pi(s;\theta):\mathbb{R}^{|S|}\rightarrow\mathbb{R}^{|U|}$ to calculate the optimal action,
and a critic network $V(s;\phi):\mathbb{R}^{|S|}\rightarrow\mathbb{R}$ to evaluate the current state $s$ under policy $\pi$.
$\theta$ and $\phi$ represent network weights in actor and critic.
$|S|$ and $|U|$ represent the dimension of state and action, respectively. Here we use $U$ rather than $A$ to represent action space and use $u$ rather that $a$ to represent an action, which matches up the expression in game theory and avoids character conflict with vehicle acceleration.

The value of critic network is equal to the sum of future rewards:

\vskip -0.1in
\begin{align}
    V(s_t;\phi) = \sum_{i=0}^\infty r(s_{t+i},u_{t+i})
    \label{ADP_value}
\end{align}
where $s_t$ is the state in time step $t$, and $r$ is the reward function.
The weights $\phi$ in critic network $V(s_t;\phi)$ is updated according to Bellman equation:

\vskip -0.1in
\begin{align}
    V(s_t;\phi) = r(s_{t},u_{t}) + V(s_{t+1};\phi)
    \label{bellman_eq}
\end{align}
and the weights $\theta$ in actor network is updated based on the reward and current critic network:

\vskip -0.1in
\begin{align}
    \pi(s_t;\theta) = \mathop{\arg \min}\limits_u \{r(s_{t},u) + V(f(s_t,u);\phi) \}
    \label{pi_update}
\end{align}
where $f:\mathbb{R}^{|S|}\times\mathbb{R}^{|U|}\rightarrow\mathbb{R}^{|S|}$ is the environment dynamic model.
Environment dynamic model could provide the next state $s_{t+1}$ given the inputs of the state and action in the current time step.

\section{Methods}
\subsection{Modeling of Non-zero-sum Game}
\label{modeling}
To model the vehicle motion more realistic, we adopt the vehicle dynamic model rather than the kinematic model.
Vehicle dynamic model is widely used in several well-known autonomous driving platforms, such as NVIDIA DRIVE Sim \cite{nvndia_sim} and Baidu Apollo \cite{fan2018baidu}.
The dynamic model makes it possible to jointly control acceleration and steering angle in continuous space.
We use the 3-DOF vehicle dynamic model derived by backward Euler [Ge \emph{et al.}, 2021] to get a stable simulation even the longitudinal vehicle speed is low.
\begin{figure}[H]
    \begin{center}
    \centerline{\includegraphics[width=0.8\columnwidth]{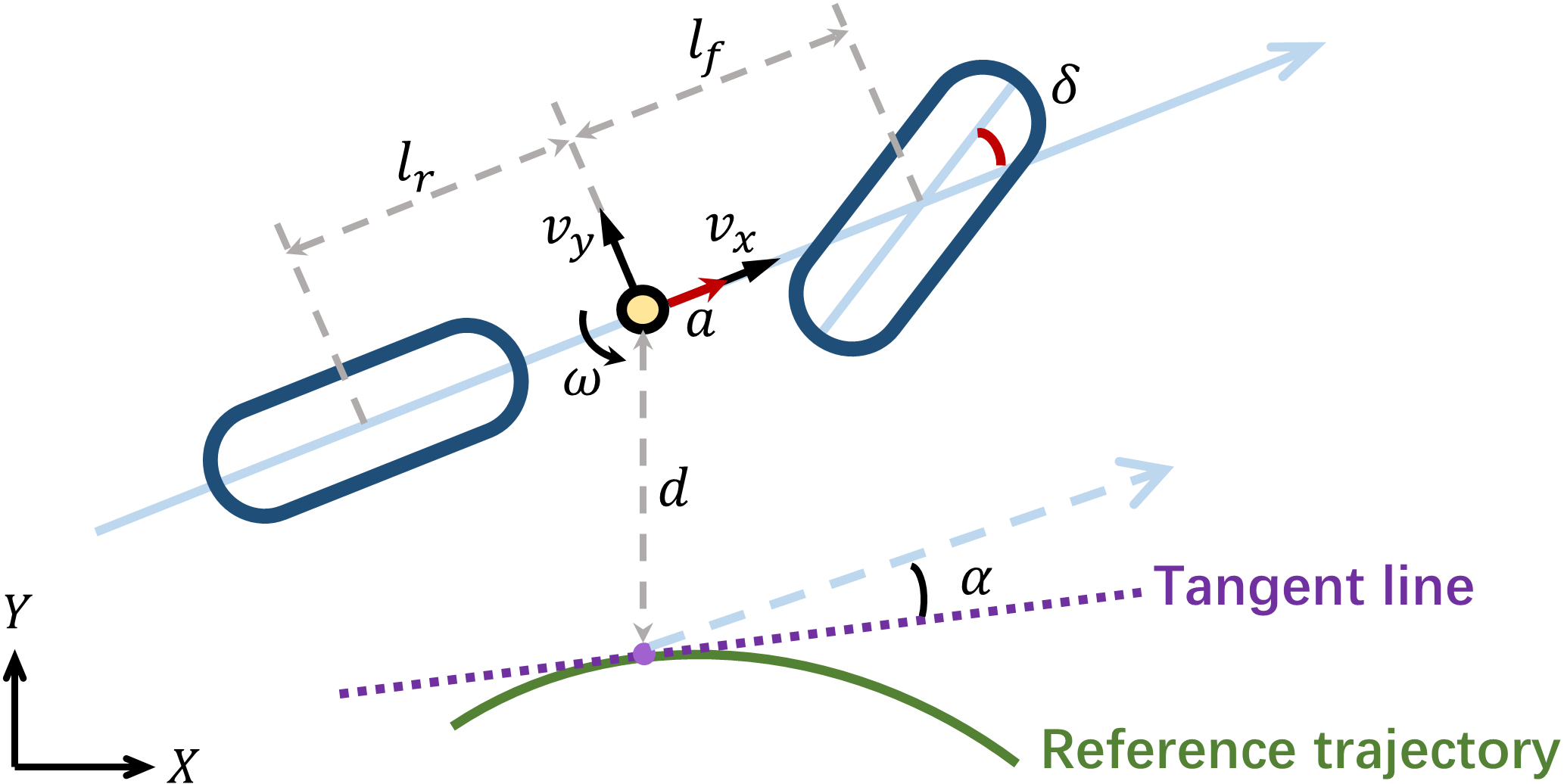}}
    \caption{\textbf{The 3-DOF vehicle dynamic model.}
             $\delta$ is the front wheel steering angle.
             $a$ is the longitudinal acceleration.
             $v_x, v_y$ are longitudinal speed and lateral speed.
             $\omega$ is the yaw rate.
             $l_f$ is the distance between the center of gravity and the front axle.
             $l_r$ is the distance between the center of gravity and the rear axle.
             $d$ is the trajectory offset from vehicle center to reference trajectory.
             $\alpha$ is the heading angle error between the vehicle heading angle and the tangent line of reference trajectory.}
    \label{vehicle_dynamic}
    \end{center}
\end{figure}
We use an intersection scenario as example, where each vehicle tracks its reference trajectory until going through the intersection.
The scenario is shown in Figure \ref{intersection_scenario}.
The vehicles' motions obey the dynamic model and they take individual actions according to different objective functions.
Objective functions includes trajectory tracking performance index, traffic efficiency index, safety performance index and comfort performance index.
This situation can be modeled as a MDP (Markov Decision Process).
We give the definations of state, action, dynamic and reward in this MDP as follows.

\textbf{State. } The state of vehicle $i$ is defined as $x_i$
\[x_i = [X_i, Y_i, d_i, \alpha_i, v_{x,i}, v_{y,i}, \omega_i]^\top,\ i\in\{1,\cdots,n\}\]
where $(X_i, Y_i)$ are the global location of vehicle $i$ and $n$ is the total number of vehicles.
Other variables are declared in Figure \ref{vehicle_dynamic}.
The system state of all vehicles is defined as $s$:
\[s = [x_1,\cdots,x_n]^\top\]
It is the combination of all vehicle states. Here we ignore timestamp $t$ for simplification.

\textbf{Action. } The action of vehicle $i$ is defined as $u_i$
\[u_i = [a_i, \delta_i]^\top\]
where $a_i$, $\delta_i$ are the longitudinal acceleration and steering angle of vehicle $i$, respectively.
Here we use $u$ to represent action rather than $a$ to match up the expression in game theory and avoid character conflict with vehicle acceleration.

\textbf{Dynamic. } After the definition of state and action, the discrete-time state transition can be written in an affine nonlinear equation (\ref{affine_dynamic}) when only controlling the steering angle:

\vskip -0.1in
\begin{align}
    s_{t+1} = f(s_t) + \sum_{i=1}^n g_i(s_t)u_i(s_t)
    \label{affine_dynamic}
\end{align}
where transform $f(s_t)$ and $g_i(s_t)$ can be found based on the vehicle dynamic model.
The exact form of $f(s_t)$ and $g_i(s_t)$ are listed in Appendix \ref{dynamic_formula}.

Furthermore, when we control acceleration and steering angle together,
the discrete-time state transition can be written in a non-affine nonlinear equation (\ref{non-affine_dynamic}):

\vskip -0.1in
\begin{align}
    s_{t+1} = f(s_t, u_1,\cdots,u_n)
    \label{non-affine_dynamic}
\end{align}
Due to the good property of the affine nonlinear system, we use equation (\ref{affine_dynamic}) rather than (\ref{non-affine_dynamic}) in the expression in section \ref{solving_ADP}.
And we simply give a convergence illustration of solving non-zero-sum game by ADP when using affine system in section \ref{solving_ADP}.
However, according to experiments in section \ref{experiments}, we found that our algorithm can also converge when using non-affine system and reach a better performance than affine system.

\textbf{Reward. } The reward $r_i$ of vehicle $i$ is defined as:

\vskip -0.1in
\begin{align}
    r_i(s, u_1,&\cdots,u_n) = \bm{\textcolor{MediumVioletRed}{c_{d} \cdot d_{i}^{2} + c_{\theta} \cdot \alpha_{i}^{2}}} \nonumber\\
                           &+ \bm{\textcolor{DodgerBlue}{c_{v} \cdot\left(v_{x, i}-v_{ref, i}\right)^{2}}} \nonumber\\
                           &+ \bm{\textcolor{OliveDrab}{c_{safe} \cdot \max \left\{0,\ 5^{2}-\left(\max_j {\rm dis}(i,j)\right)^2\right\}}} \nonumber\\
                           &+ \bm{\textcolor{DarkOrange}{c_{\delta} \cdot \delta_{i}^{2}+c_{a} \cdot a_{i}^{2}}}
    \label{reward}
\end{align}
where $v_{ref, i}$ is the reference longitudinal speed of vehicle $i$,
${\rm dis}(i,j)$ is the Euclidean distance of vehile $i$ and $j$.
And $c_*$ are hyper-parameter constants.
Other variables are already declared in this section.

The {red} part in equation (\ref{reward}) indicates \bm{\textcolor{MediumVioletRed}{trajectory tracking performance}}.
The {blud} part indicates \bm{\textcolor{DodgerBlue}{traffic efficiency}}.
The {green} part indicates \bm{\textcolor{OliveDrab}{safety performance}}.
The {orange} part indicates \bm{\textcolor{DarkOrange}{comfort performance}}.
All of the parts except safety index are in quadratic form. There is a convergence guarantee of using ADP to solve non-zero-sum game when reward is in quadratic, as we illustrated in section \ref{solving_ADP}.
According to experiments in section \ref{experiments}, we found that our algorithm can also converge when the safety index is not quadratic.

\subsection{Solving Coupled HJB Equation}
\label{solving_ADP}
According to equation (\ref{ADP_value}),
the value function is the sum of future rewards.
By the Bellman's principle of optimality, the optimal value function of vehicle $i$ should be

\vskip -0.1in
\begin{align}
    V_{i}^{*}(s_t)=\min_{u_{i}}\left\{ r_{i}\left(s_t, u_{1}, \ldots, u_n\right)+V_{i}^{*}(s_{t+1}) \right\}
    \label{optimal_value}
\end{align}
Next state $s_{t+1}$ is derived based on the environment dynamic model.
We assume that equation (\ref{optimal_value}) is continuous and differentiable, the environment dynamic model is affine, the reward is in quadratic form. Then

\vskip -0.1in
\begin{align}
    \frac{\partial V_{i}^{*}(s_t)}{\partial u_{i}} = 2 R_{ii} u_{i}+g_{i}^\top(s_t) \frac{\partial V_{i}^{*}(s_{t+1})}{\partial s_{t+1}}
    \label{dev_value}
\end{align}
where $R_{ii}$ is the diagonal matrix containing the weights in quadratic reward function, and $g_i$ is the matrix in equation (\ref{affine_dynamic}).
According to stationarity conditions of optimization,
which is the first-order optimality condition, we can get the necessary condition of Nash equilibrium:

\vskip -0.1in
\begin{align}
    \frac{\partial V_{i}^{*}(x)}{\partial u_{i}}=0,\ \forall i
    \label{optimal_condition}
\end{align}
Take equation (\ref{optimal_condition}) into equation (\ref{dev_value}), get

\vskip -0.1in
\begin{align}
    u_{i}^{*}=-\frac{1}{2} R_{ii}^{-1} g_{i}^\top(s_t) \nabla V_{i}^{*}(s_{t+1}),\ \forall i
    \label{control_law}
\end{align}
where $\nabla V_{i}^{*}(s_{t+1})=\frac{\partial V_{i}^{*}(s_{t+1})}{\partial s_{t+1}}$.

Take equation (\ref{control_law}) into equation (\ref{optimal_value}), get the coupled HJB (Hamilton-Jacobi-Bellman) equations

\vskip -0.1in
\begin{align}
    \label{HJB_eq}
    0&=Q_{i}(s_t)+V_{i}^{*}\left(s_{t+1}\right)-V_{i}^{*}(s_t) \\
    &+\frac{1}{4} \sum_{j=1}^{n}\left(\nabla V_{j}^{*}\left(s_{t+1}\right)\right)^\top g_{j}(s_t) R_{jj}^{-1} g_{j}^\top(s_t) \nabla V_{j}^{*}\left(s_{t+1}\right), \forall i \nonumber
\end{align}
Note that there are $n$ equations in (\ref{HJB_eq}) and each equation contains $n$ value functions $V_1^*, \ldots, V_n^*$, therefore it is called \textit{coupled} HJB equations.
In equation (\ref{HJB_eq}), the $n$ value functions $V_{i}^{*}$ are the only unknown quantities.
In simple tasks, we can directly solve equation (\ref{HJB_eq}) to get $V_{i}^{*}$ in a closed-form solution.
Then the Nash equilibrium control strategy can be obtained by taking $V_{i}^{*}$ into the equation (\ref{control_law}).
However, due to the severe nonlinearity of the multi-vehicle driving problem, equation (\ref{HJB_eq}) can only be solved by numerical methods.

For the nonlinear non-zero-sum game with known dynamic, ADP can be used to solve it \cite{vamvoudakis2011multi,zhang2016discrete,zhang2012near,li2013adaptive}.
Convergence guarantees are provided in \cite{vamvoudakis2011multi,zhang2016discrete} when the dynamic system is affine and reward function is quadratic.
Noteworthily, these works are all verified in simple tasks, e.g. the system with just $2$ or $3$ dimension state.
And some works just use polynomials as the function approximator due to the straightforward system dynamic.
In our scenario, we extend to a high-dimensional system with at least $14$ dimension state and use deep neural networks as the function approximator.

\begin{figure}[h]
    \begin{center}
    \centerline{\includegraphics[width=0.8\columnwidth]{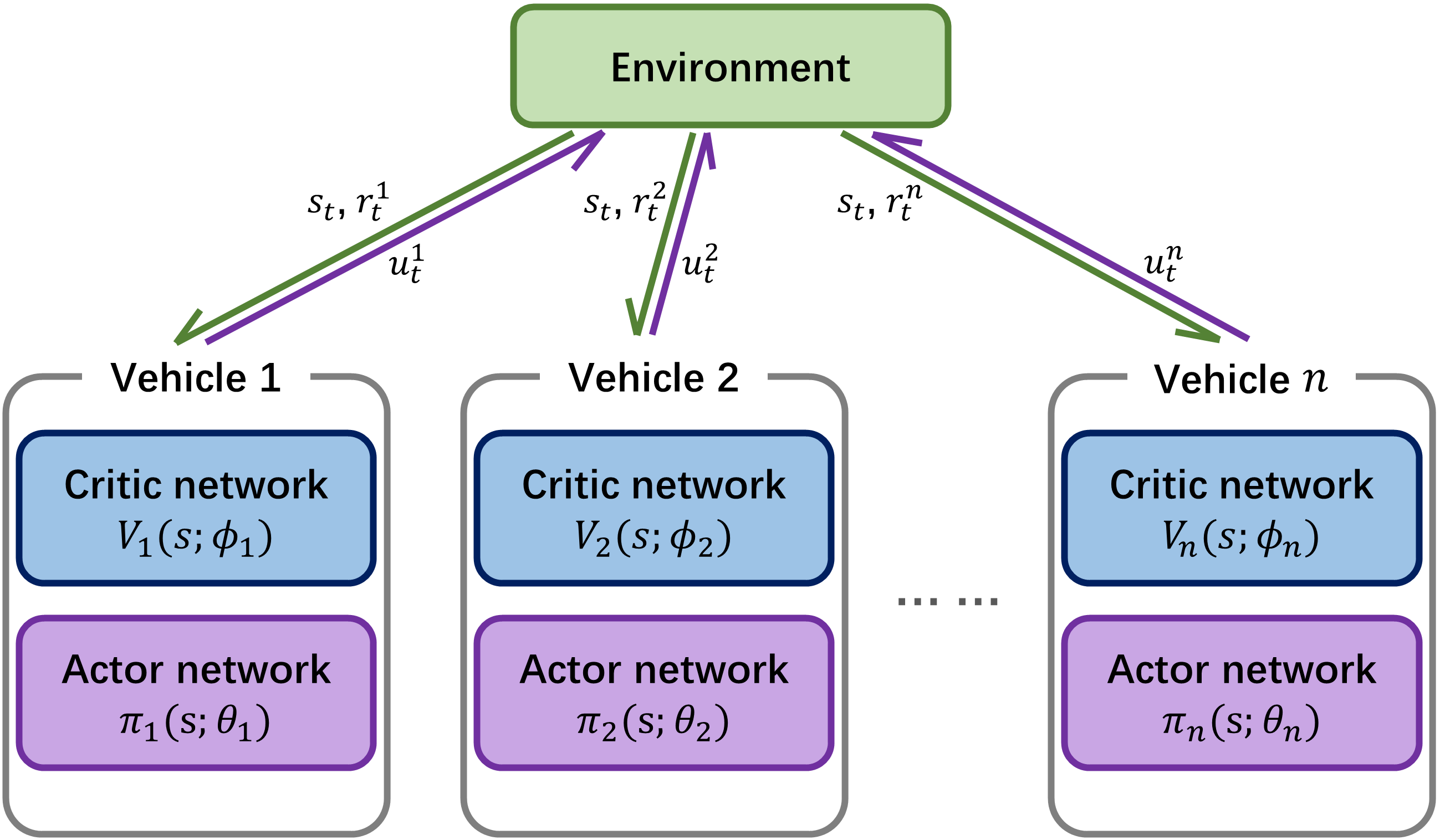}}
    \caption{\textbf{Interactive flow chart of non-zero-sum driving control.}
    To represent time step and vehicle index at the same time when writing reward $r$ and action $u$, we place the time step and vehicle index in subscript and superscript, respectively.}
    \label{alg_1_flow}
    \end{center}
\end{figure}

To increase the training stability and convergence speed, we also adopt \textit{Experience Replay} \cite{mnih2013playing}. 
Finally, the method is described in Algorithm \ref{alg_1}.
The interactive flow chart is shown in Figure \ref{alg_1_flow}.
\begin{algorithm}[h]
    \caption{Solve Nash Equilibrium by ADP}
    \label{alg_1}
 \begin{algorithmic}
    \STATE {\bfseries Initialize:} $V_i(s;\phi_i)$ and $\pi_i(s;\theta_i),\ \forall i$
    \REPEAT
        \STATE Use $\pi_i(s;\theta_i)$ run an episode and store the state trajectory ($s_0,\ldots,s_T$) into replay buffer. 
        \STATE Sample a batch of $s_t$ from replay buffer.
        \STATE Calculate actions, rewards and next states:
            \vskip -0.25in
            \begin{align*}
                u_i &= \pi_i(s_t;\theta_i)\\
                r_i &= r_i(s_{t},u_1,\ldots,u_n)\\
                s_{t+1} &= f(s_{t}, u_1,\ldots,u_n)
            \end{align*}
        \STATE Calculate critic loss and actor loss according to equations (\ref{bellman_eq}) and (\ref{pi_update}):
            \vskip -0.25in
            \begin{align*}
                {\rm c\_loss}_{i} &= \left(r_i + V_i(s_{t+1};\phi_i) - V_i(s_t;\phi_i)\right)^2\\
                {\rm a\_loss}_{i} &= r_i + V_i(s_{t+1};\phi_i)
            \end{align*}
        \STATE Update $V_i(s;\phi_i)$ and $\pi_i(s;\theta_i)$ by gradient descent:
            \vskip -0.2in
            \begin{align*}
                \phi_i &\leftarrow \phi_i - \eta \frac{\partial {\rm c\_loss}_i}{\partial \phi_i} \\
                \theta_i &\leftarrow \theta_i - \eta \frac{\partial {\rm a\_loss}_i}{\partial \theta_i}
            \end{align*}
    \UNTIL{ $V_i(s;\phi_i)$ and $\pi_i(s;\theta_i)$ converge, $\forall i$. }
 \end{algorithmic}
\end{algorithm}
\subsection{Training by Self-play}
\label{self_play_design}
Self-play is an important trick for training the reinforcement learning agent.
The agent will be trained by regarding itself as competitors.
In this way, the intelligence and training efficiency will be improved.
For example, AlphaGo \cite{silver2016mastering}, AlphaGo Zero \cite{silver2017mastering} and DouZero \cite{zha2021douzero} are all trained in self-play.

In Algorithm \ref{alg_1}, there are $n$ critic networks $V_i(s;\phi_i)$ and $n$ actor networks $\pi_i(s;\theta_i)$ corresponding to $n$ vehicles.
When the number of vehicles $n$ is big enough, the size of training parameter space becomes too large.
To overcome this issue, we design a parameter-sharing mechanism to decrease the number of parameters and make it possible to self-play training.

\begin{figure}[h]
    \begin{center}
    \centerline{\includegraphics[width=\columnwidth]{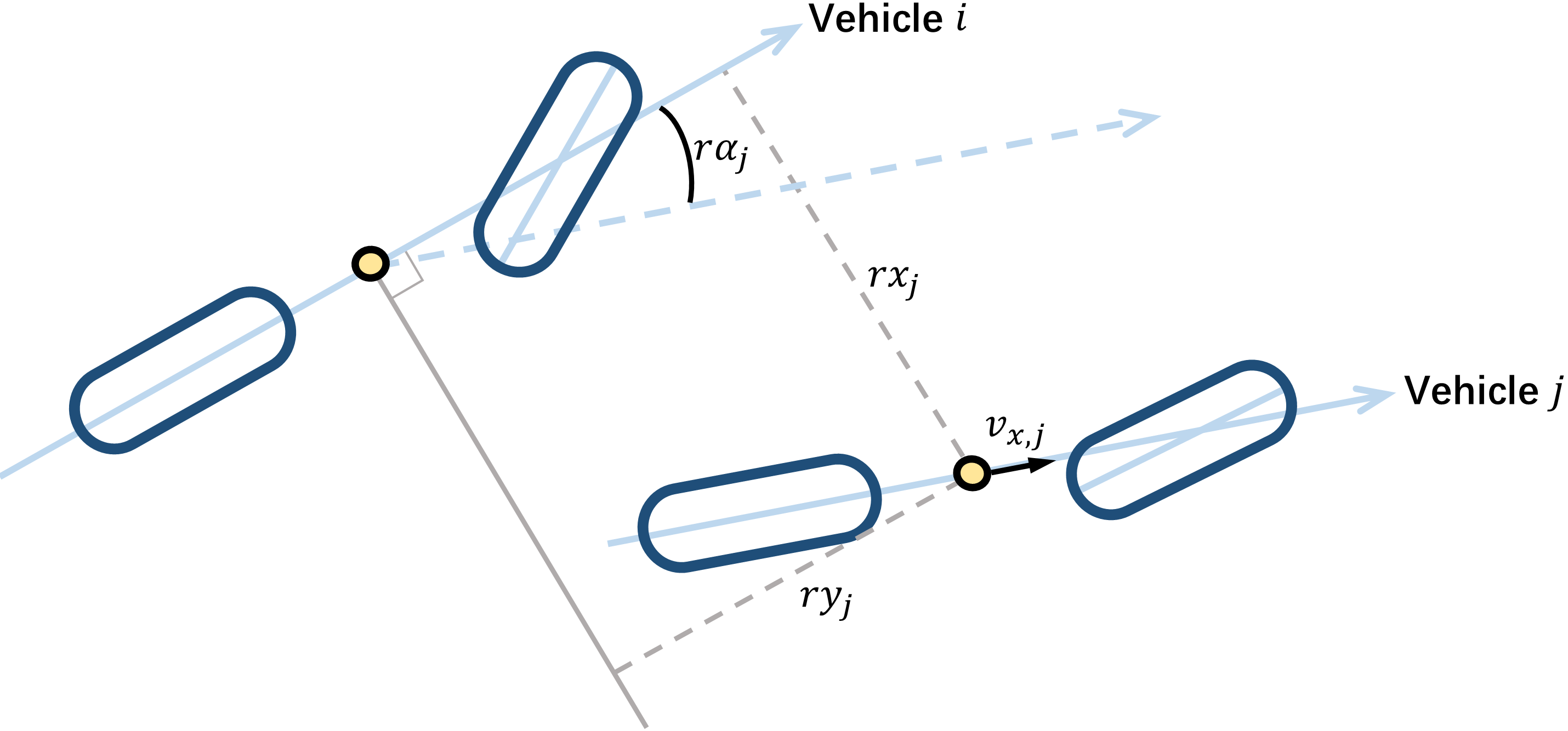}}
    \caption{\textbf{Relative state} of surrounding vehicle $j$ in the observation view of vehicle $i$.}
    \label{relative_state}
    \end{center}
\end{figure}

Specifically, we first transform the state $s_t$ from the global observation view to a local observation view.
When vehicle $i$ perceives environment state, the surrounding vehicles' state will be observed in vehicle $i$'s view.
Let's define the relative state of surrounding vehicle $j$ in the observation view of vehicle $i$ as
\[rx_j = [rx_j, ry_j, r\alpha_j, v_{x,j}]^\top\]
where $(rx_j, ry_j)$ is the relative coordinate in the view of vehicle $i$,
$r\alpha_j$ is the relative heading angle,
$v_{x,j}$ is the longitudinal speed of vehicle $j$.
The variables explanation is shown in Figure \ref{relative_state}.
Then, the overall state in the view of vehicle $i$ is defined as
\[s_i = [d_i, \alpha_i, v_{x,i}, v_{y,i}, \omega_i, rx_1^\top, \ldots, rx_{n-1}^\top]^\top\]
It contains the dynamic state of vehicle $i$ and the relative states of all surrounding vehicles.

\begin{figure}[h]
    \begin{center}
    \centerline{\includegraphics[width=\columnwidth]{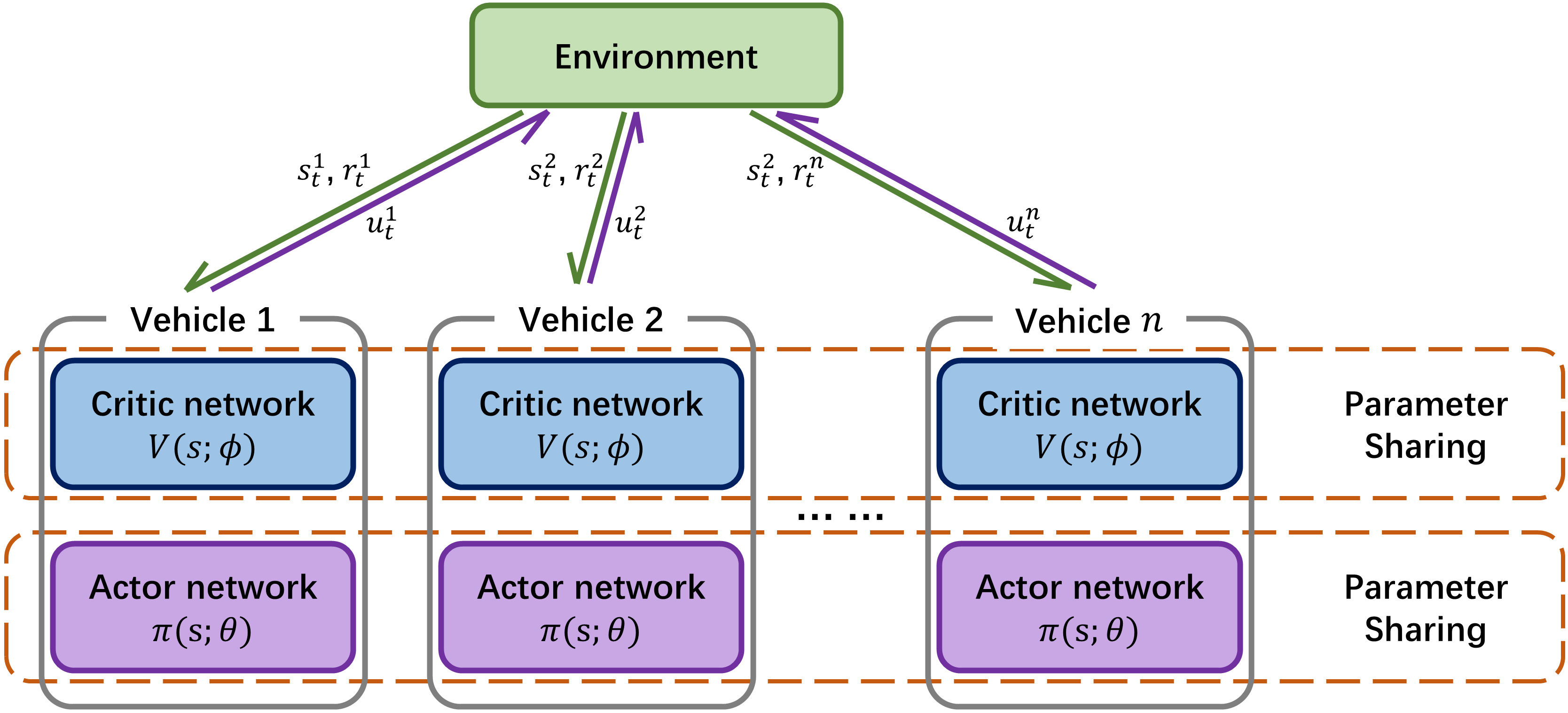}}
    \caption{\textbf{Interactive flow chart of non-zero-sum driving control when using self-play training.}
    To represent time step and vehicle index at the same time when writing reward $r$ and action $u$, we place the time step and vehicle index in subscript and superscript, respectively.}
    \label{alg_2_flow}
    \end{center}
\end{figure}

In this way, the observed state of each vehicle has a unified form, which makes it possible to adopt the parameter-sharing mechanism.
There will be only one critic network $V(s;\phi)$ with parameter $\phi$ suitable for each vehicle,
and only one actor network $\pi(s;\theta)$ with parameter $\theta$ suitable for each vehicle.
And we suppose that the hyper-parameters in the reward function of each vehicle are the same.
After that, the Algorithm \ref{alg_1} changes from mutual-play to self-play, which is shown in Algorithm \ref{alg_2}.
The interactive flow chart is shown in Figure \ref{alg_2_flow}.

\begin{algorithm}[H]
    \caption{Self-play ADP for Solving Nash Equilibrium}
    \label{alg_2}
 \begin{algorithmic}
    \STATE {\bfseries Initialize:} $V(s;\phi)$ and $\pi(s;\theta)$
    \REPEAT
        \STATE Use $\pi(s;\theta)$ to run an episode and store the trajectory of unified-form state ($s_0,\ldots,s_T$) into replay buffer. 
        \STATE Sample a batch of $s_t$ from replay buffer.
        \STATE Calculate actions, rewards and next states:
            \vskip -0.25in
            \begin{align*}
                u &= \pi(s_t;\theta)\\
                r &= r(s_{t},u_1,\ldots,u_n)\\
                s_{t+1} &= f(s_{t}, u_1,\ldots,u_n)
            \end{align*}
        \STATE Calculate critic loss and actor loss according to equations (\ref{bellman_eq}) and (\ref{pi_update}):
            \vskip -0.25in
            \begin{align*}
                \rm c\_loss &= \left(r + V(s_{t+1};\phi) - V(s_t;\phi)\right)^2\\
                \rm a\_loss &= r + V(s_{t+1};\phi)
            \end{align*}
        \STATE Update $V(s;\phi)$ and $\pi(s;\theta)$ by gradient descent:
            \vskip -0.2in
            \begin{align*}
                \phi &\leftarrow \phi - \eta \frac{\partial {\rm c\_loss}}{\partial \phi} \\
                \theta &\leftarrow \theta - \eta \frac{\partial {\rm a\_loss}}{\partial \theta}
            \end{align*}
    \UNTIL{ $V(s;\phi)$ and $\pi(s;\theta)$ converge. }
 \end{algorithmic}
\end{algorithm}

\section{Experiments}
\label{experiments}
Intersection are known as a high collision risk scene, where about 61\% of side collisions happened \cite{accidents}.
There will be more interactions between vehicles if intersection has no traffic light, which increases the driving dangerousness.
Therefore, we choose intersection without traffic light as the standard scene for experiments.

We describe three experiments in this section.
The first experiment aims to achieve intersection passing via lateral control.
The second experiment aims to achieve intersection passing via both lateral and longitudinal control.
We demonstrate two vehicles scenario as an example in the first and second experiments.
In the third experiment, four vehicles are going through the intersection,
which could verify the ability to scale up thanks to self-play training.
The neural networks used in the three experiments are the same, whose structure is shown in Appendix \ref{net_structure}.

\begin{figure*}[b]
    \begin{center}
        \subfigure[Vehicles trajectory]{\includegraphics[width=0.675\columnwidth]{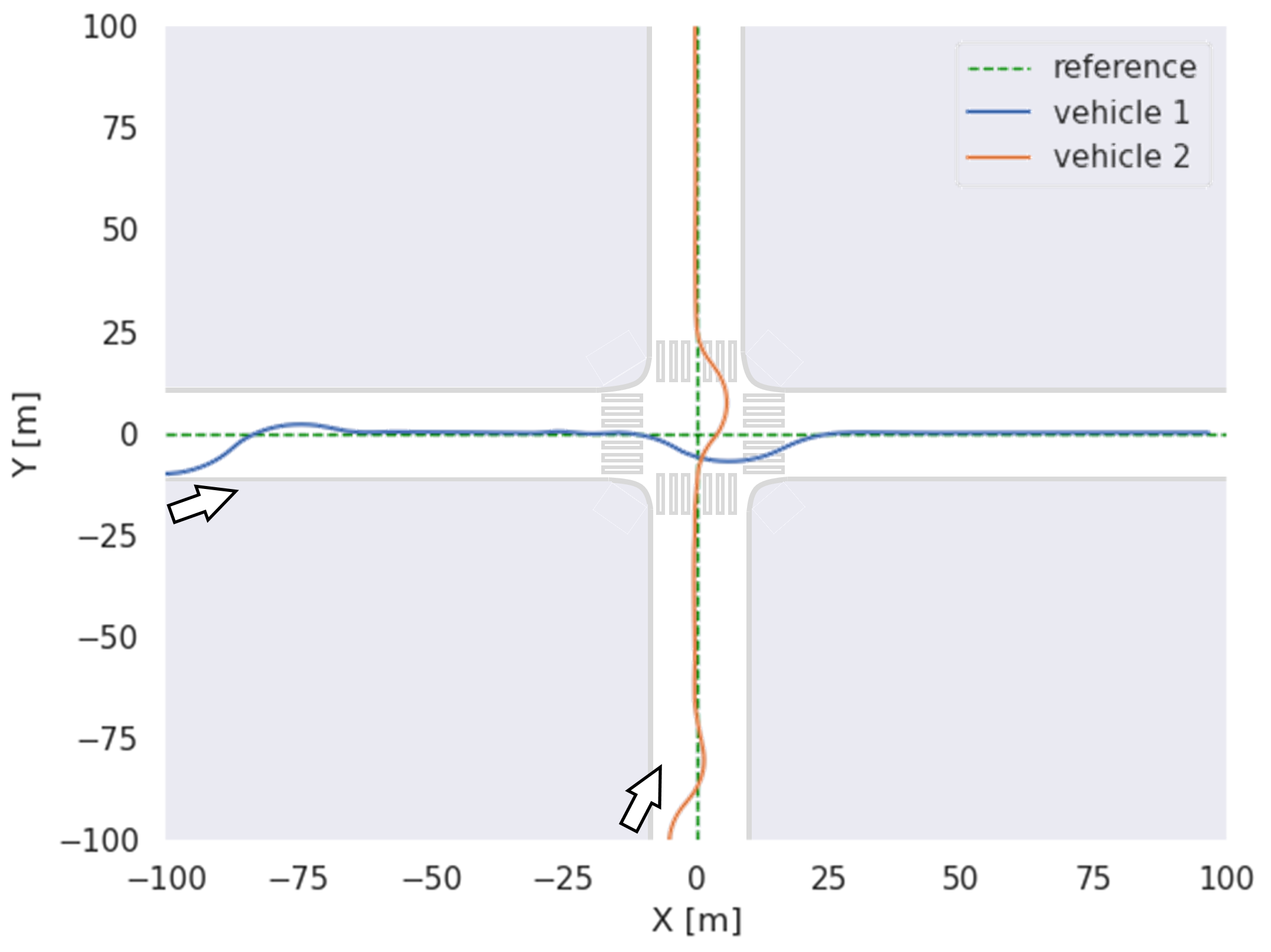}}
        \subfigure[Distance between vehicles]{\includegraphics[width=0.67\columnwidth]{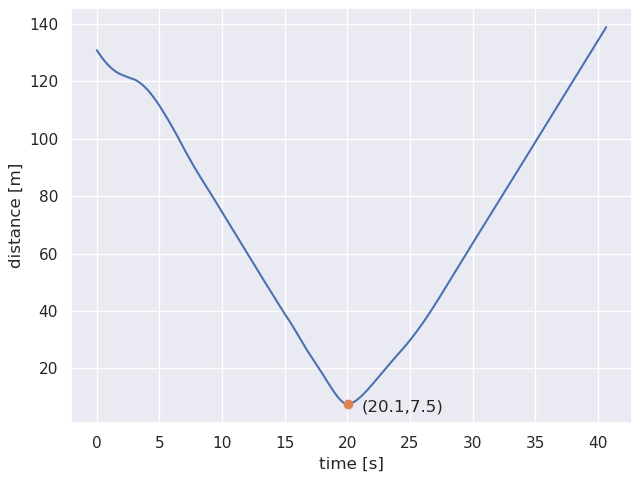}}
        \subfigure[Reward curve]{\includegraphics[width=0.67\columnwidth]{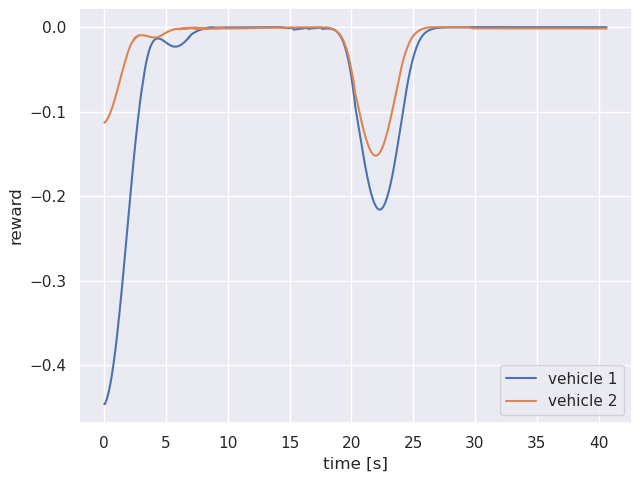}}
        \vskip -0.1in
        \caption{\textbf{Lateral control result}.}
        \label{exp1}
    \end{center}
\end{figure*}

\textbf{Lateral Control. } In the first experiment, we describe a two-vehicle driving scenario at a non-signalized intersection.
Vehicles can only control the steering angle to track their reference trajectories and avoid collisions.
The longitudinal speeds are set as a constant $5\ \rm{m/s}$.

There is a non-signalized intersection formed by two reference trajectories marked in green dash line,
which is shown in Figure \ref{exp1}{\rm (a)}.
The vechiles aims to track the reference trajecotries to pass the intersection.
The two reference trajectories meet at point $(0,0)$.
The first vehicle drives from left to right, whose 
real trajectory is marked in blue.
The second vehicle drives from down to up, whose real trajectory is marked in orange.
At the beginning, vehicle 1 and vehicle 2 are located at $(-100,-10)$ and $(-5,-100)$,
which means they are $10$m and $5$m away from their reference trajectories, respectively.
When the simulation starts, two vehicles track their reference trajectories successfully by controlling the steering angle.
When they approach the center of intersection, both of them try to be the first one to pass.
However, because the arrival time of vehicle11 is slightly later than vehicle 2,
vehicle 1 can only turn the steering wheel right and give the way for passing the intersection as soon as possible.
Figure \ref{exp1}{\rm (b)} implies that the nearest distance between vehicle 1 and 2 is $7.5\ \rm m$,
which is bigger than the accepted safety distance $5\ \rm m$ set in equation \ref{reward}.
Figure \ref{exp1}{\rm (c)} implies the reward of vehicle 1 is less than vehicle 2 when they meet at the center of intersection because vehicle 1 gives the way.

\textbf{Lateral and Longitudinal Control. } In the second experiment, we describe a two-vehicle driving scenario at a non-signalized intersection.
Vehicles can control both steering angle and acceleration to track their reference trajectories, track their reference speed and avoid collisions.

\begin{figure*}[t]
    \begin{center}
        \subfigure[Vehicles trajectory]{\includegraphics[width=0.67\columnwidth]{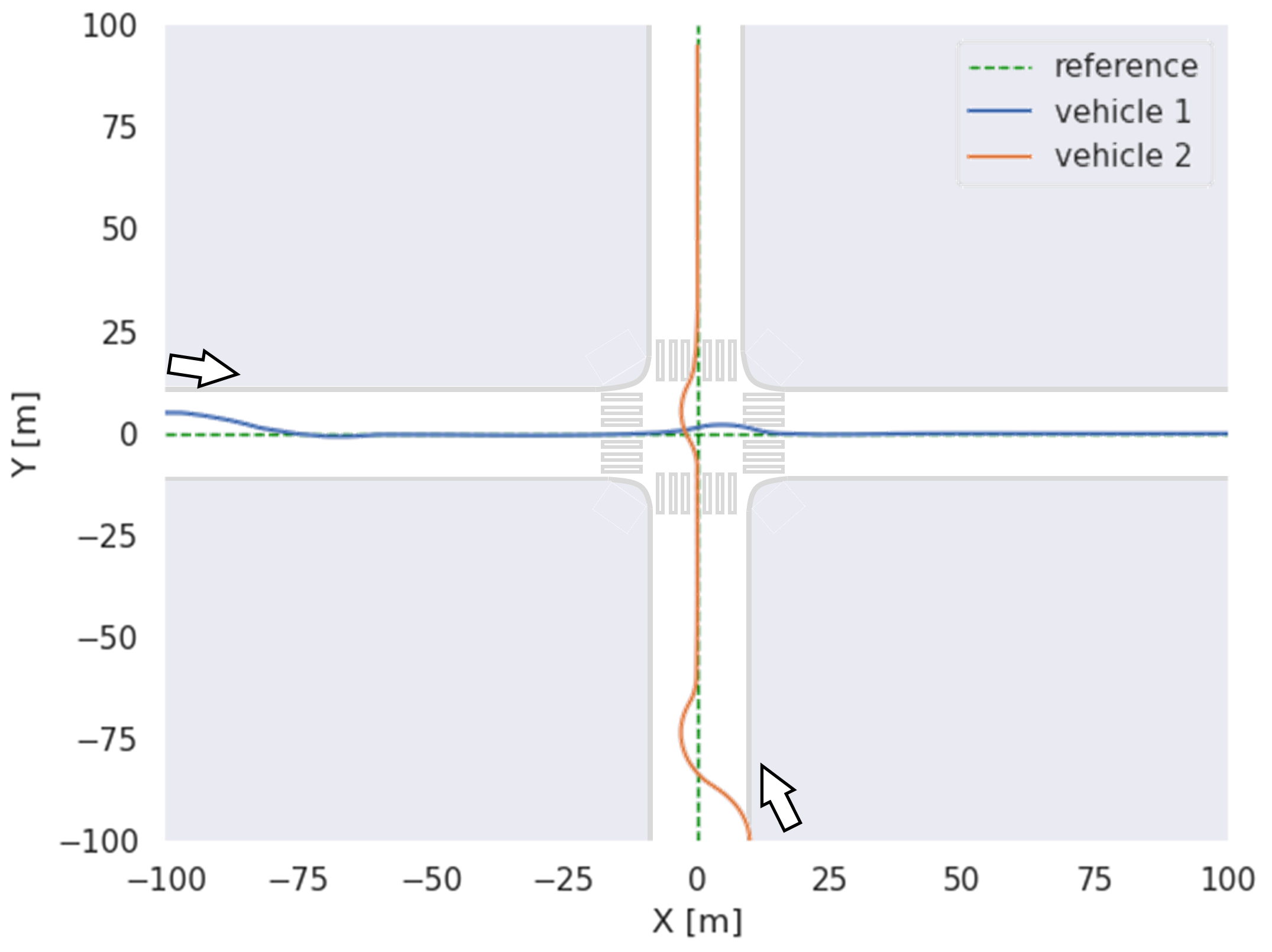}}
        \subfigure[Longitudinal speed]{\includegraphics[width=0.67\columnwidth]{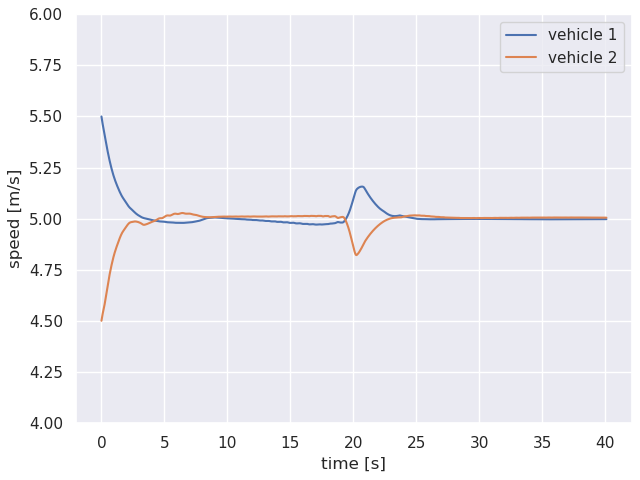}}
        \subfigure[Distance between vehicles]{\includegraphics[width=0.67\columnwidth]{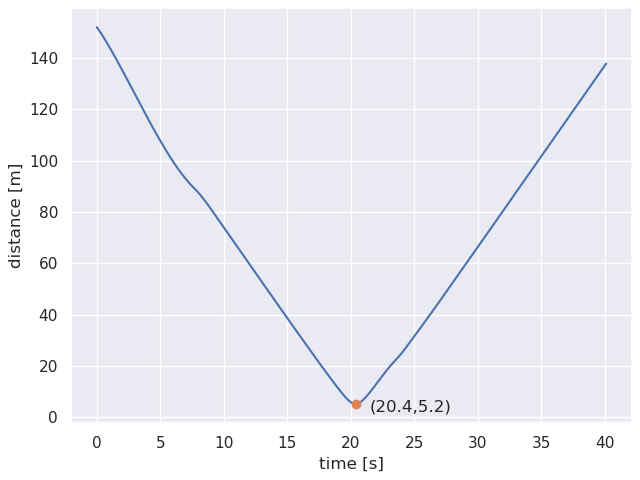}}
        \vskip -0.1in
        \caption{\textbf{Lateral and longitudinal control result}.}
        \label{exp2}
    \end{center}
\end{figure*}

In Figure \ref{exp2}{\rm (a)}, reference trajectories are marked in green dash line, who meet at point $(0,0)$.
The vehicle 1 in blue drives from left to right.
The vehicle 2 in orange drives from down to up.
At the beginning, vehicle 1 and vehicle 2 are located at $(-100,5)$ and $(10,-100)$,
which means they are $5\ \rm m$ and $10\ \rm m$ away from their reference trajectories, respectively.
As shown in Figure \ref{exp2}{\rm (b)}, their initial longitudinal speeds are $5.5\ \rm{m/s}$ and $4.5\ \rm{m/s}$, respectively.
When the simulation starts, two vehicles track their reference trajectories and reference speeds successfully by controlling steering angle and acceleration, as shown in Figure \ref{exp2}{\rm (a)(b)} and \ref{exp2}{\rm (b)}.
When they approach the intersection, both of them try to be the first one to pass.
However, because the arrival time of vehicle 2 is slightly later than vehicle 1,
vehicle 2 can only turn the steering wheel left and decelerate to give way
while vehicle 1 turn the steering wheel left and accelerates to rush.
Figure \ref{exp2}{\rm (c)} implies that the nearest distance between them is $5.2\ \rm m$,
which is bigger than the accepted safety distance $5\ \rm m$ set in equation \ref{reward}.

Comparing Figure \ref{exp1}{\rm (a)} and Figure \ref{exp2}{\rm (a)},
the trajectory offsets at the intersection center decrease from about $10\ \rm m$ to about $3\ \rm m$ after considering longitudinal control.
It is because the vehicle can avoid collision not only by turning the steering wheel,
but also by adjusting the acceleration.
Therefore, the lateral and longitudinal control model is more reasonable than the lateral control model although no convergence proof.

\begin{figure*}[htbp]
    \begin{center}
        \subfigure[Vehicles trajectory]{\includegraphics[width=0.678\columnwidth]{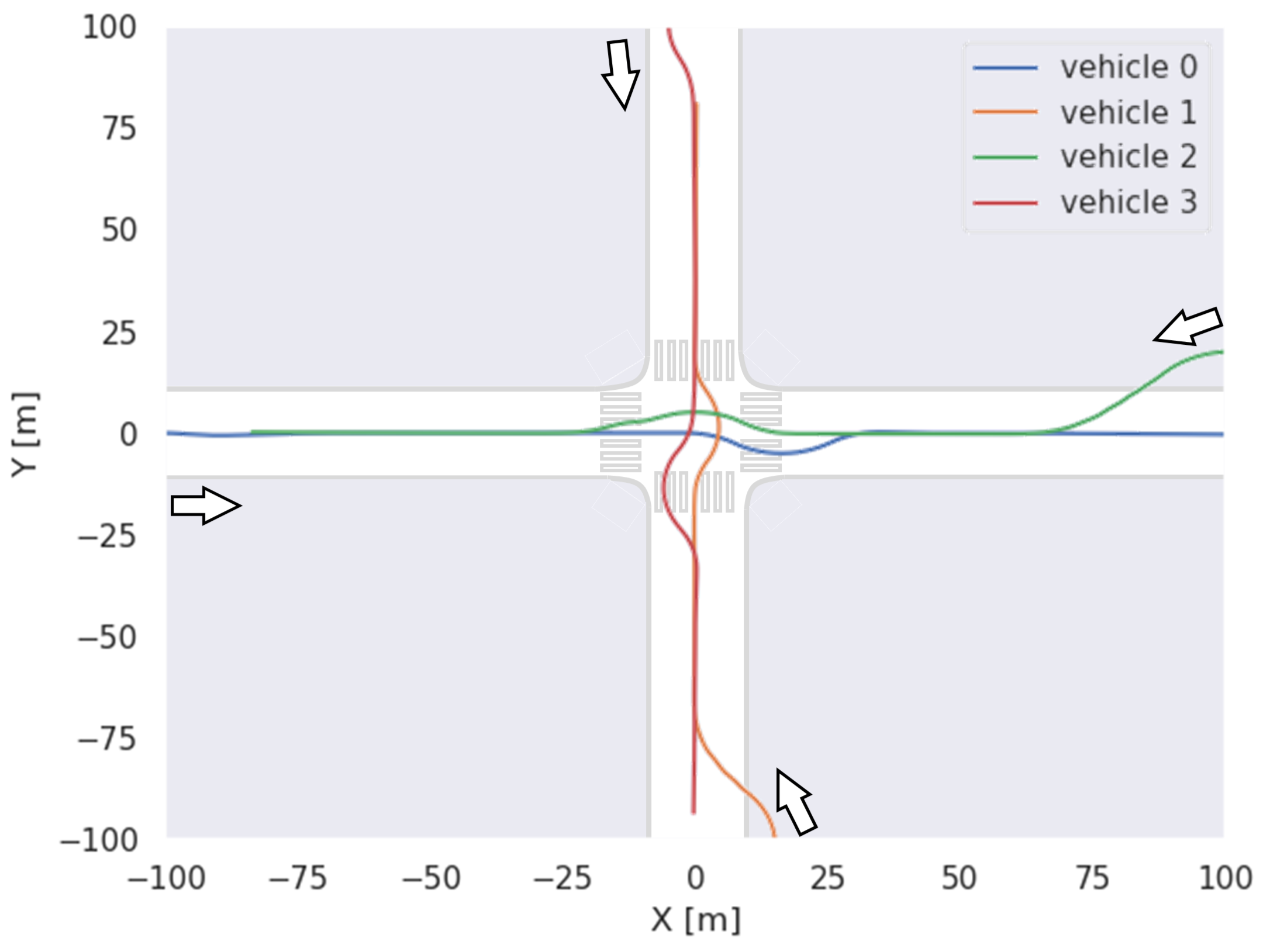}}
        \subfigure[Longitudinal speed]{\includegraphics[width=0.67\columnwidth]{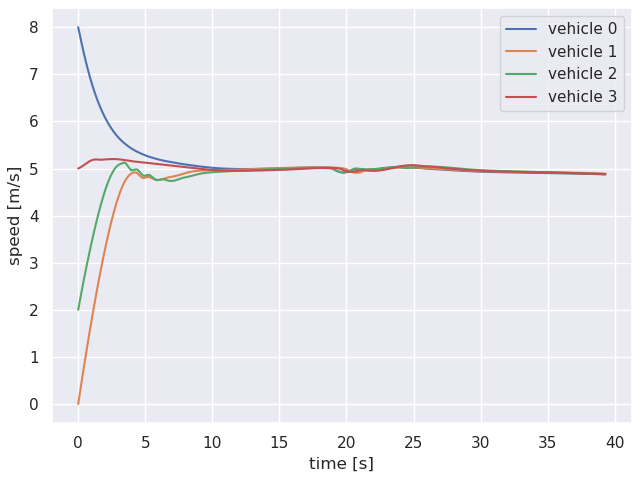}}
        \subfigure[Reward curve]{\includegraphics[width=0.67\columnwidth]{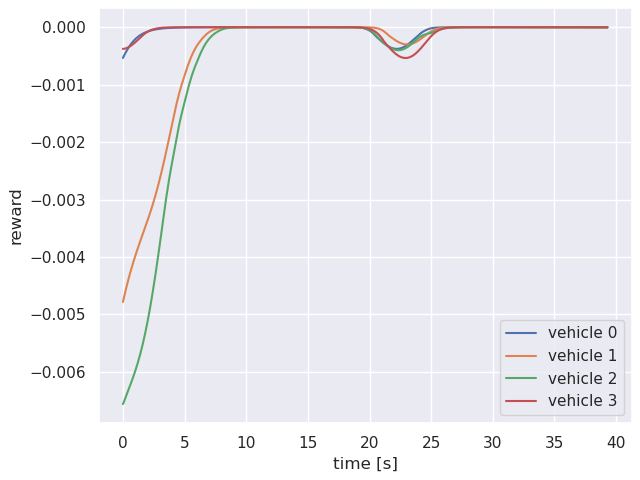}}
        \vskip -0.1in
        \caption{\textbf{Self-play training result}.}
        \label{exp3}
    \end{center}
\end{figure*}

\textbf{Self-play Training. } In the third experiment, we describe a four-vehicle driving scenario at a non-signalized intersection via lateral and longitudinal control.
Parameter sharing mechanism is adopted and Algorithm \ref{alg_2} is used to train the agents.
This experiment aims to show the effectiveness of self-play training designed in section \ref{self_play_design}.

As shown in Figure \ref{exp3}{\rm (a)}, the initial trajectory offsets of these four vehicles range from $0\ \rm m$ to $25\ \rm m$.
And their initial speeds range from $0\ \rm{m/s}$ to $7\ \rm{m/s}$, which is shown in Figure \ref{exp3}{\rm (b)}.
Figure \ref{exp3}{\rm (c)} shows the rewards of vehicles during the simulation.
When the simulation starts, all of the vehicles track their reference trajectories and reference speeds successfully by controlling steering angle and acceleration.
When they approach the intersection center, they show the behavior of overtaking and passing in Figure \ref{exp3}{\rm (a)}.
It implies the agents have learned to overtake and pass by itself, even though we do not set any traffic rules for overtaking and passing.
The simulation video can be seen here \footnote{{\rm https://github.com/jerry99s/NonZeroSum\_Driving}}. 

\section{Conclusion}
Multi-vehicles driving on the road is a complex interactive game problem.
In this paper, we construct the multi-vehicle driving scenario as a non-zero-sum game and propose a novel game control framework, which considers prediction, decision and control as a whole.
The mutual influence of interactions between vehicles is considered in this framework because decisions are made by Nash equilibrium strategies.
We provide an effective way to solve the Nash equilibrium driving strategies based on model-based reinforcement learning.
Then we introduce experience replay and self-play tricks for scaling up to a larger agent number.
Experiments show that the Nash equilibrium driving strategies trained by our algorithm could drive perfectly. Furthermore, vehicles have learned to overtake and pass without prior knowledge.

In the future, we should consider more scenarios besides intersections.
Hyper-parameters in reward function should also be adaptable to surrounding cars suitable for different driving styles.

\appendix

\bibliographystyle{named}
\bibliography{references}

\clearpage
\section*{Appendix}
\section{Vehicle Dynamic Model}
\label{dynamic_formula}
\subsection{Lateral Control Model}
\label{lateral_model}
Assume the state of a vehicle is defined as $x$
\[x = [X, Y, d, \alpha, v_{x}, v_{y}, \omega]^\top\]
The action of a vehicle is defined as $u$
\[u = [a, \delta]^\top\]
The variables' meaning and units are listed in Table \ref{table_dynamic}.

\begin{table}[h]
    \caption{Description of the variables in vehicle dynamic model.}
    \label{table_dynamic}
    \begin{center}
    \begin{small}
    \begin{tabular}{cll}
    \toprule
    Variable & Description & Unit \\
    \midrule
    $X$     & global coordinate $X$ of vehicle      & $\rm m$ \\
    $Y$     & global coordinate $Y$ of vehicle      & $\rm m$ \\
    $d$     & trajectory offset                     & $\rm m$ \\
    $\alpha$& heading angle error                   & $\rm rad$ \\
    $v_x$   & longitudinal speed                    & $\rm m/s$ \\
    $v_y$   & lateral speed                           & $\rm m/s$ \\
    $\omega$& yaw rate                              & $\rm rad/s^2$ \\
    $a$     & longitudinal acceleration             & $\rm m/s^2$\\
    $\delta$& front wheel steering angle            & $\rm rad$\\
    \bottomrule
    \end{tabular}
    \end{small}
    \end{center}
\end{table}
The state differential equation is

\vskip -0.1in
\begin{align}
    \dot{x} =
    \begin{bmatrix}
        v_x \cos \beta - v_y \sin \beta \\
        v_x \sin \beta + v_y \cos \beta \\
        \dot{d} \\
        0 \\
        \omega \\
        -v_x\omega + \frac{F_f\cos \delta + F_r}{m} \\
        \frac{l_f F_f\cos\delta - l_r F_r}{I_z}
    \end{bmatrix}
    \label{differential_eq}
\end{align}
If the vehicle drives parallel to the $X$-axis, $\beta=\alpha$ and $\dot{d}=\dot{Y}$.
If the vehicle drives parallel to the $Y$-axis, $\beta=\alpha+\pi/2$ and $\dot{d}=\dot{X}$.
$m$ is the vehicle mass.
$l_f$ is the distance between the center of gravity and the front axle.
$l_r$ is the distance between the center of gravity and the rear axle.

The lateral force on the front wheel is

\vskip -0.1in
\begin{align*}
    F_f &= C_f(\theta_{vf} - \delta) \\
        &\approx C_f(\tan \theta_{vf} - \delta) \\
        &= C_f\left(\frac{v_y+l_f \omega}{v_x} - \delta\right)
\end{align*}
and the lateral force on the rear wheel is

\vskip -0.1in
\begin{align*}
    F_r &= C_r\theta_{vr} \\
        &\approx C_r\tan \theta_{vr} \\
        &= C_r\frac{v_y-l_r \omega}{v_x}
\end{align*}
where $C_f$, $C_r$ are the cornering stiffness and $\theta_{vf}$, $\theta_{vr}$ are wheel slip angles.

Assume $\delta$ is small enough, then
\[\sin\delta\approx0,\  \cos\delta\approx1\]
Take the above equations into equation (\ref{differential_eq}), get

\vskip -0.1in
\begin{align}
    \dot{x} =
    \begin{bmatrix}
        v_x \cos \beta - v_y \sin \beta \\
        v_x \sin \beta + v_y \cos \beta \\
        \dot{d} \\
        0\\
        \omega \\
        -v_x\omega + \frac{C_f(v_y+l_f \omega)+C_r(v_y-l_r \omega)}{m\cdot v_x} - \frac{C_f}{m}\delta \\
        \frac{l_f C_f(v_y+l_f \omega)-l_r C_r(v_y-l_r \omega)}{I_z\cdot v_x} - \frac{l_f C_f}{I_z}\delta \\ 
    \end{bmatrix}
    \label{differential_eq2}
\end{align}
Then, we can write equation (\ref{differential_eq2}) into affine nonlinear form:
\[\dot{x} = f(x) + g(x)u \]
where $f(x)=\begin{bmatrix}
                    v_x \cos \beta - v_y \sin \beta \\
                    v_x \sin \beta + v_y \cos \beta \\
                    \dot{d} \\
                    0\\
                    \omega \\
                    -v_x\omega + \frac{C_f(v_y+l_f \omega)+C_r(v_y-l_r \omega)}{m\cdot v_x}\\
                    \frac{l_f C_f(v_y+l_f \omega)-l_r C_r(v_y-l_r \omega)}{I_z\cdot v_x}\\ 
            \end{bmatrix}$\\
and $g(x)=[0,0,0,0,0,- \frac{C_f}{m},- \frac{l_f C_f}{I_z}]^\top$.

We set parameters according to cars in the real world.
The parameters setting is listed in Table \ref{para_set}.
\begin{table}[h]
    \caption{Parameters setting in the vehicle dynamic model.}
    \label{para_set}
    \begin{center}
    \begin{small}
    \begin{tabular}{ccl}
    \toprule
    Parameter & Value & Unit \\
    \midrule
    $C_f$       & $-88000$        & $\rm N/rad$ \\
    $C_r$       & $-94000$        & $\rm N/rad$ \\
    $l_r$       & $1.14$          & $\rm m$ \\
    $l_r$       & $1.4$           & $\rm m$ \\
    $m$         & $1500$          & $\rm kg$ \\
    $I_z$       & $2420$          & $\rm kg\cdot m^2$ \\
    $v_x$       & $10$            & $\rm m/s$ \\
    $\delta$    & $\left[-0.35,0.35\right]$  & $\rm rad$\\
    $\Delta t$  & $0.05$          & $\rm second$\\
    \bottomrule
    \end{tabular}
    \end{small}
    \end{center}
    \vskip -0.1in
\end{table}

\subsection{Lateral and Longitudinal Control Model}
The lateral control model will become unstable when the longitudinal speed is low.
Therefore, we use a low-speed stable lateral and longitudinal control model [Ge \emph{et al.}, 2021].

Assume the state of a vehicle is defined as $x$
\[x = [X, Y, d, \alpha, v_{x}, v_{y}, \omega]^\top\]
The action of a vehicle is defined as $u$
\[u = [a, \delta]^\top\]
The discrete state transform equation is:

\[    x' =
    \begin{bmatrix}
        X+\Delta t(v_x\cos \beta-v_y \sin \beta)\\
        Y+\Delta t(v_x\sin \beta+v_y \cos \beta)\\
        d+\Delta d \\
        v_x +\Delta t\cdot a\\
        \theta +\Delta t\cdot \omega \\
        v_y + \frac{m\cdot v_x v_y +\Delta t(l_f C_f -l_r C_r)\omega - \Delta t \cdot C_f v_x \delta - \Delta t\cdot m\cdot v_x^2\omega}{m\cdot v_x-\Delta t(C_f+C_r)} \\
        \frac{I_z v_x \omega+\Delta t(l_f C_f - l_r C_r)v_y-\Delta t\cdot l_f C_f v_x \delta}{I_z v_x - \Delta t(l_f^2 C_f+l_r^2 C_r)}
    \end{bmatrix}
\]
where $x'$ is the vehicle state in the next time step.
The parameters setting is listed in Table \ref{para_set}, which is the same as the lateral control model in Appendix \ref{lateral_model}.

\section{Neural Network Structure}
\label{net_structure}
\begin{figure}[h]
    \begin{center}
        \subfigure[Critic network]{\includegraphics[width=0.4\columnwidth]{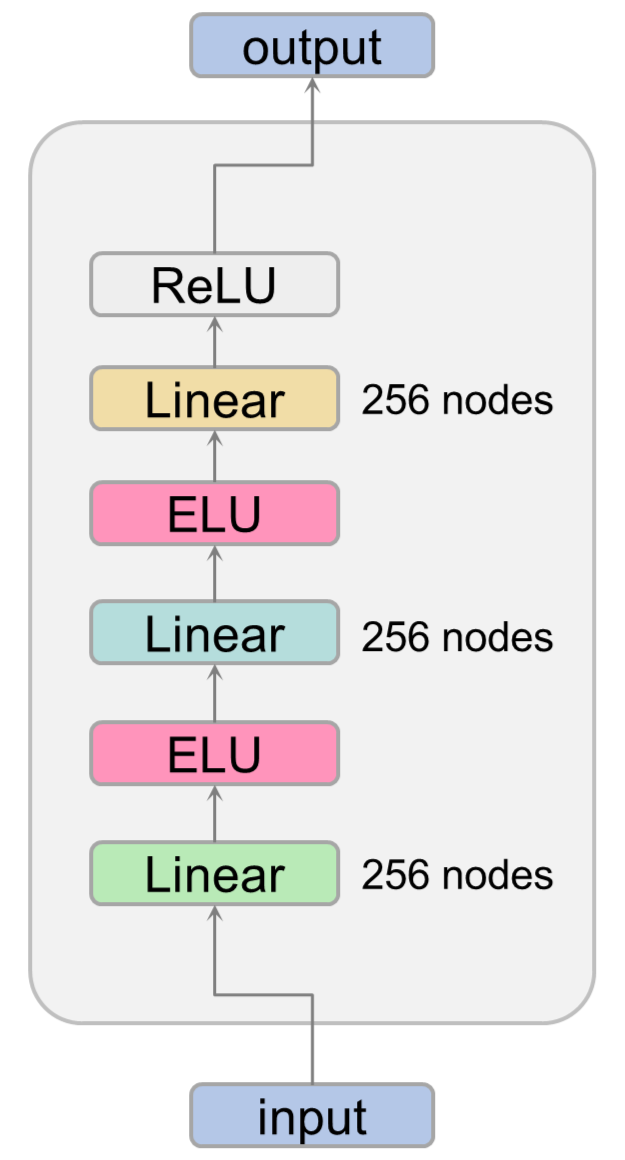}}
        \subfigure[Actor network]{\includegraphics[width=0.4\columnwidth]{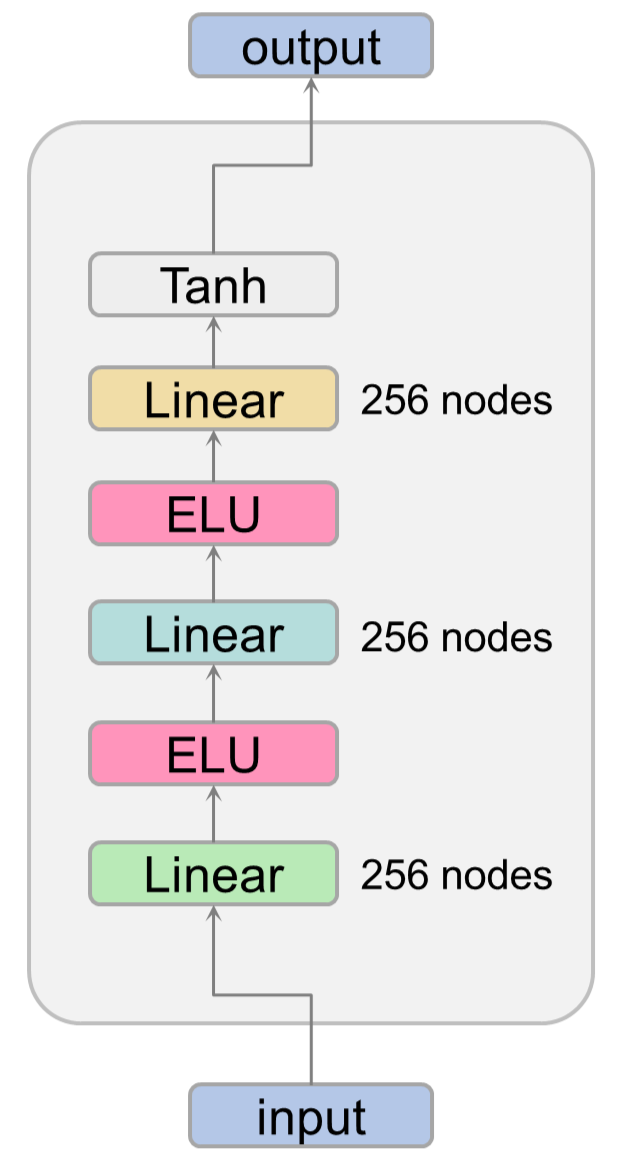}}
        \vskip -0.1in
        \caption{Neural network structures.}
        \label{fig:net_struct}
    \end{center}
\end{figure}

In experiments mentioned in section \ref{experiments},
we use the neural networks shown in Figure \ref{fig:net_struct}.
The $\tanh$ in actor network restricts the output in $[-1,1]$.
The final action output should be the actor network output multiplied by the action amplitude.

\end{document}